# Cross-lingual Dysarthria Severity Classification for English, Korean, and Tamil

Eun Jung Yeo* Kwanghee Choi† Sunhee Kim‡ and Minhwa Chung§
* Seoul National University, Seoul, Korea
E-mail: ej.yeo@snu.ac.kr Tel/Fax: +82-2-8806162
† Sogang University, Seoul, Korea
E-mail: juice500@sogang.ac.kr Tel/Fax: +82-2-8806162
‡ Seoul National University, Seoul, Korea
E-mail: sunhkim@snu.ac.kr Tel/Fax: +82-2-8807693
§ Seoul National University, Seoul, Korea
E-mail: mchung@snu.ac.kr Tel/Fax: +82-2-8809195

***Abstract*—Data scarcity hinders research on dysarthria severity classification due to the limited size of datasets. While the cross-lingual approach has been applied to alleviate the problem, the roles of language-specific features have been underestimated. This paper proposes a cross-lingual classification method for English, Korean, and Tamil, which employs both language-independent features and language-unique features. First, we extract thirty-nine features from diverse speech dimensions such as voice quality, pronunciation, and prosody. Second, feature selections are applied to identify the optimal feature set for each language. A set of shared features and a set of distinctive features are distinguished by comparing the feature selection results of the three languages. Lastly, automatic severity classification is performed, utilizing the two feature sets. Notably, the proposed method removes different features by languages to prevent the negative effect of unique features for other languages. Accordingly, eXtreme Gradient Boosting (XGBoost) algorithm is employed for classification, due to its strength in imputing missing data. In order to validate the effectiveness of our proposed method, two baseline experiments are conducted: experiments using the intersection set of mono-lingual feature sets (Intersection) and experiments using the union set of mono-lingual feature sets (Union). According to the experimental results, our method achieves better performance with a 67.14% F1 score, compared to 64.52% for the Intersection experiment and 66.74% for the Union experiment. Further, the proposed method attains better performances than mono-lingual classifications for all three languages, achieving 17.67%, 2.28%, 7.79% relative percentage increases for English, Korean, and Tamil, respectively. The result specifies that commonly shared features and language-specific features must be considered separately for cross-language dysarthria severity classification.**

## I. INTRODUCTION

Dysarthria is a group of motor speech disorders resulting from a neuromuscular control disturbance, affecting multiple speech dimensions such as respiration, phonation, articulation, resonance, and prosody [1]. This results in degraded speech intelligibility, repeated communication failures, and ultimately poor quality of life. Consequently, accurate and reliable intelligibility assessment is essential in tracking patients' conditions and treatments' effectiveness. Perceptual evaluation of intelligibility is often conducted by speech pathologists, which is often immensely labor- and time-consuming. Hence, the automatic classification that highly correlates with the experts will have great potential for aiding clinicians in diagnosis and speech therapy [2][3].

There are two main approaches for the automatic assessment of dysarthria. The first approach is to propose a novel neural network that takes raw speech signals as input [4][5][6][7]. However, dysarthria speech datasets are often too small, prohibiting the use of large neural networks that are often highly sample-inefficient. Further, its black-box nature fundamentally obstructs its interpretability, which clinicians often crave.

Earlier studies have explored the efficacy of various features on dysarthria severity classification, albeit most are mono-lingual. Cross-lingual approach is beneficial in terms of increasing the number of training samples. Further, the approach has the advantage of using natural speech compared to other data augmentation methods, such as speech synthesis [8], a transformation of healthy speech to dysarthric speech [9][10], perturbations on both healthy and dysarthric speech [11], and multi-resolution feature extraction [12].

Cross-lingual studies have generally focused on identifying features that can discriminate severity levels of dysarthria, while independent of the native language. Experiments on Korean and American English pointed out smaller vowel space for dysarthric speakers compared to healthy controls [13]. Experiments on Castilian Spanish, Colombian Spanish, and Czech found plosives, vowels, and fricatives to be relevant acoustic segments for dysarthria detection [14]. Research on dysarthria detection argued that the number of inter-word pauses per minute and median Harmonics to Noise Ratio play significant roles for both Czech and American English [15][16].

Nonetheless, the previous literature claimed the intelligibility of dysarthria to be significantly sensitive to native languages. First, despite the findings of compressed vowel space in dysarthric speakers for both Korean and American English, multiple regression models of speech intelligibility displayed different results, even with the same predictor variables and speakers matched on speech intelligibility [13]. Experiment using GMM-UBM classifiers reached accuracy between 85% and 94% for within-language experiments, but accuracy between 75% and 82% were achieved for experiments with combined corpora [14]. Dysarthria detection attained 73% and 72% classification accuracy for each Czech and US English datasets, but the performance dropped to 67% when the datasets were combined [16][16]. The results imply that even with seemingly common characteristics, naive cross-lingual training will yield suboptimal performances, requiring careful design of the feature set. Previous findings further suggest that features should be distinguished into language-independent and language-sensitive features, and both information must be considered for cross-lingual classification. Hence, we aim to present an effective consolidation of the two feature sets.

This paper proposes a cross-lingual dysarthria severity classification method that separately employs shared and distinctive features of languages. Especially, the proposed method removes different features by languages, in order to prevent the intervening effect of unique features of other languages. Accordingly, eXtreme Gradient Boosting (XGBoost) algorithm is utilized for classification, due to its strength in imputing missing data. We conduct cross-lingual experiments on languages with different prosodic systems, such as English, Korean, and Tamil. Three languages are chosen to demonstrate the effectiveness of our method, robust to diverse languages with different characteristics. First, features from diverse speech dimensions, including voice quality, pronunciation (phoneme correctness, vowel distortion), and prosody (speech rate, pitch, loudness, rhythm), are extracted. Second, feature selections are applied to find the optimal feature set for each language. Then, language-independent and language-unique features are distinguished by comparing the feature selection results. Lastly, automatic severity classifications that separately utilize the two feature sets are conducted.

The rest of the paper is organized as follows: Section II introduces the proposed method, and Section III describes acoustic features used for classification. Section IV presents three datasets: TORGO for English, QoLT database for Korean, and SSNCE database for Tamil. Finally, Section V describes the experimental settings and reports classification results, which is followed by a conclusion in Section VI.

## II. Methods

Section II introduces the proposed cross-lingual classification method consisting of three steps: feature extraction, feature selection, and classification[1]. We utilize eXtreme Gradient Boosting (XGBoost) algorithm [17], which has been applied to various medical data and reported robust performances [18][19][20][21]. XGBoost is a distributed gradient-boosted decision tree (GBDT) machine learning library. GBDT is a decision tree ensemble learning algorithm, which is similar to random forest, but different in terms of the strategies for combining the trees: while the bagging technique is used to minimize the variance and overfitting for random forests, boosting helps minimize the bias and underfitting for GBDT. Here, boosting refers to improving a single weak model by ensembling the number of other weak models. Further, error residuals are used for the iterative generation of weak models and training ensembles of the trees [22]. XGBoost is recognized not only for its ability to capture non-linear interactions between the features and the target variable but handling the missing values. For the missing values, the algorithm assigns the default direction, by computing loss for both directions and selecting the best of the two [23].

### A. Feature extraction

A total of 39 acoustic features from diverse speech dimensions are extracted. Feature list is designed to reflect common perceptual characteristics from various speech dimensions: voice quality, pronunciation (phoneme correctness, vowel distortion), and prosody (speech rate, pitch, loudness, rhythm). MFCCs are additionally extracted for baseline experiments, to check the efficacy of the 39 acoustic features in severity classification. Further explanations on the features are introduced in Section III and Table IV.

### B. Feature selection

To efficiently select the optimal feature sets for each language, feature selection is applied to each language feature set, by using XGBoost algorithm. A set of language-specific features and a set of language-independent features are determined by comparing the feature selection results of the three languages. MFCCs, as baseline feature set, are excluded from feature selection.

### C. Classification

Automatic severity classification is conducted with XGBoost algorithm, which employs both language-independent features and language-specific features. As described in Table I, the proposed model utilizes shared features across languages while language-specific features are used only for the corresponding target language. In detail, feature A is shared between all three languages: English, Korean, and Tamil. Feature B is shared between Korean and Tamil, while feature C is a unique feature for Tamil. Accordingly, unselected feature values, such as feature B from English, and feature C from English and Korean are dropped for cross-lingual classification. This way, the method benefits from the increased number of training samples from the shared features and exploits language-unique features without interference by unrelated features from other languages.

TABLE I: Tabular for the Proposed experiment

| Language | A | B | C |
|---|---|---|---|
| English | O | N/A | N/A |
| Korean | O | O | N/A |
| Tamil | O | O | O |

[1]https://github.com/eunjung31/crossLingual-XGBoost.git

To validate the effectiveness of our proposed method, we test different combinations of language-independent and language-specific feature set: Intersection and Union. **Intersection** experiment exploits language-independent features only, which is determined as commonly selected features by all three languages. **Union** experiment, on the other hand, uses all features that are selected from other languages. Table II demonstrates the tabular for intersection and union experiments. Descriptions of the experiments are described below.

For a language $l$ and all the feature set $U$, let language-specific feature set $F_l \subset U$ and language-specific dataset with selected features $F$ be $D^l_F$.

1) (intersection) $\bigcup_{l \in \{\texttt{en},\texttt{ko},\texttt{ta}\}} D^l_{F_{\texttt{en}} \cap F_{\texttt{ko}} \cap F_{\texttt{ta}}}$
2) (union) $\bigcup_{l \in \{\texttt{en},\texttt{ko},\texttt{ta}\}} D^l_{F_{\texttt{en}} \cup F_{\texttt{ko}} \cup F_{\texttt{ta}}}$
3) (proposed) $\bigcup_{l \in \{\texttt{ko},\texttt{en},\texttt{ta}\}} D^l_{F_l}$.

TABLE II: Tabular for Intersection and Union experiments

(a) Intersection

| Language | A | B | C |
|---|---|---|---|
| English | O | N/A | N/A |
| Korean | O | N/A | N/A |
| Tamil | O | N/A | N/A |

(b) Union

| Language | A | B | C |
|---|---|---|---|
| English | O | O | O |
| Korean | O | O | O |
| Tamil | O | O | O |

## III. Acoustic features

### A. MFCCs

MFCCs describe the overall spectral envelope shape. Mean and standard deviation of 12-dim MFCCs and log energy set are extracted by using librosa [24].

### B. Voice quality

For voice quality features, seven acoustic features are extracted: jitter, shimmer, Harmonics to Noise Ratio (HNR), Period Perturbation Quotient (PPQ), Amplitude Perturbation Quotient (APQ), number of voice breaks, and degree of voice breaks. All features are commonly used for the diagnosis of phonation quality. Dysarthric speakers are commonly reported to have higher values for all features except HNR, compared to healthy speakers [25][26].

Jitter and shimmer each measures the variations of fundamental frequency (F0) and amplitude, respectively. Equation (1) and Equation (2) each presents the absolute jitter and absolute shimmer. PPQ and APQ are the proposed values to exclude the natural change of F0 over time from jitter and shimmer [27]. Equation (3) and Equation (4) are the equations for absolute PPQ and APQ. $T_i$ is the duration of the $i$th interval, $A_i$ the amplitude of $i$th interval, and $N$ the number of intervals. For normalization, this study utilizes relative values, where absolute jitter and PPQ are divided by the average duration of intervals, and absolute shimmer and APQ are divided by the average amplitude of intervals. Lastly, as shown in Equation (5), HNR represents the energy of the periodic portion ($E_p$) divided by the energy of the noise energy ($E_n$).

$$\text{absJitter} = \frac{1}{N-1}\sum_{i=2}^{N}|T_i - T_{i-1}| \tag{1}$$

$$\text{absShimmer} = \frac{1}{N-1}\sum_{i=2}^{N}|A_i - A_{i-1}| \tag{2}$$

$$\text{absPPQ} = \sum_{i=3}^{N-2}\frac{|T_i - \frac{T_{i-2}+T_{i-1}+T_i+T_{i+1}+T_{I+2}}{5}|}{N-4} \tag{3}$$

$$\text{absAPQ} = \sum_{i=3}^{N-2}\frac{|A_i - \frac{A_{i-2}+A_{i-1}+A_i+A_{i+1}+A_{i+2}}{5}|}{N-4} \tag{4}$$

$$\text{HNR (db)} = 10\log\left(\frac{E_p}{E_n}\right) \tag{5}$$

Voice breaks are defined as inter-pulse intervals longer than 17.86ms, calculated by the distance of successive glottal pulses of 1.25 divided by the pitch floor of 70 Hz. The degree of voice breaks refers the total duration of voice breaks divided by the total duration. Features are extracted with the voice report function from Praat [28].

### C. Pronunciation - Phoneme correctness

As for phoneme correctness features, Percentage of Correct Consonants (PCC), Percentage of Correct Vowels (PCV), and Percentage of Total correct Phonemes (PCT) are extracted. PCC, PCV, and PCT are each refer to the number of correctly pronounced consonants, vowels, and phonemes (consonants and vowels) divided by the number of target consonants, vowels, and phonemes in an utterance. With limitations in movements of the articulators, dysarthric speakers generally have poor pronunciation accuracy, hence lower phoneme correctness [29]. Features are extracted by using publicly released XLSR-53 wav2vec models fine-tuned with each language [30][31][32]. The decoded and canonical phoneme sequences are aligned, as described in Table III. In detail, the example sentence consists of five consonants (HH, L, L, R, L) and eight vowels (IY, W, IH, AH, AW, AH, EH, AY). Two consonants (L, L) and five vowels (IY, W, AH, AW, AY) match with the canonical phoneme sequence. Consequently PCC results in 2/5*100 = 40.00%, PCV 5/8*100 = 62.50%, and PCT 7/15*100 = 53.85%.

TABLE III: Alignment for phoneme correctness extraction

| Canonical phoneme sequence |
|---|
| HH IY W IH L AH L AW AH * R EH L AY |
| **Decoded phoneme sequence** |
| SH IY W AO L AH L AW AE N L IY * AY |

### D. Pronunciation - Vowel space features

Five features related to vowel space are extracted: Vowel Space Area (triangular VSA, quadrilateral VSA), Formant Centralized Ratio (FCR), Vowel Articulatory Index (VAI), and F2-ratio. VSA refers to the area of the F1-F2 vowel diagram, determined by the first and second formants of the corner vowels. FCR and VAI are the features that are proposed to be more sensitive to intelligibility classification than VSA. Lastly, F2-ratio is the ratio of the second formants between the representative front vowel and the back vowel. Because people with dysarthria have restricted motor controls, their vowel space gets small as dysarthria severity level gets worse [33]. For extraction, we first create Textgrids in phoneme alignments. While Montreal Forced Aligner [34] is utilized for English and Korean, time-aligned phonetic transcription provided by the database is converted into Textgrids for Tamil. Secondly, formants are extracted from the center of the vowels by using Praat. Lastly, extracted formants are fed into the equations stated from Equation (6) to Equation (10). For materials that do not contain the required corner vowels, we interpolate the formants with the average formant values of each speaker.

TABLE IV: Acoustic features

| Category | | Features |
|---|---|---|
| Voice quality | | jitter, shimmer, PPQ, APQ, HNR, number of voice breaks, degree of voice breaks |
| Pronunciation | Phoneme correctness | PCC, PCV, PCP |
| | Vowel distortion | triangular VSA, quadrilateral VSA, FCR, VAI, F2-Ratio |
| Prosody | Speech rate | speaking rate, articulation rate, number of pauses, pause duration, phone ratio |
| | Pitch | mean, std, min, max, range of F0 |
| | Loudness | mean, std, min, max, range of energy |
| | Rhythm | %V, deltaV/C, VarcoV/C, rPVIV/C, nPVIV/C |

$$\text{VSA } \triangle = \frac{1}{2}|F1_i\left(F2_a - F2_u\right) + F1_a\left(F2_u - F2_i\right) + F1_u\left(F2_i - F2_a\right)| \quad (6)$$

$$\text{VSA } \square = \frac{1}{2}|(F2_iF1_{ae}+F2_{ae}F1_a + F2_aF1_u + F2_uF1_i) - (F1_iF2_{ae}+F1_{ae}F2_a + F1_aF2_u + F1_uF2_i)| \quad (7)$$

$$\text{FCR} = \frac{F2_u + F2_a + F1_i + F1_u}{F2_i + F1_a} \quad (8)$$

$$\text{VAI} = \frac{F2_i + F1_a}{F2_u + F2_a + F1_i + F1_u} \quad (9)$$

$$\text{F2-Ratio} = \frac{F2_i}{F2_u} \quad (10)$$

### *E. Prosody - Speech rate features*

For speech rate features, we include speaking rate, articulation rate, number of pauses, pause duration, and phone ratio. Speaking rate is defined as the number of produced syllables divided by the utterance length. Similarly, articulation rate is the number of syllables produced in utterance length excluding pauses. As for pauses-related features, intervals longer than 0.1 ms of silence are considered as pauses. Phone ratio is computed by dividing the total length of the non-silent intervals by total duration. All speech rate features are extracted by using Parselmouth [35].

### *F. Prosody - Pitch and Loudness features*

Mean, standard deviation, minimum, maximum, and range of the F0 and energy each stands for pitch and loudness features, respectively. Extraction is employed using Parselmouth, with intervals of 0 Hz or 0 dB excluded.

### *G. Prosody - Rhythm features*

%V, deltas, Varcos, rPVIs, nPVIs are used as rhythm features [36][26]. %V refers to the proportion of vocalic intervals, while deltas are standard deviations of vocalic or consonantal interval duration. As presented in Equation (11) and Equation (12), VarcoV and VarcoC are the normalized values, where delta values are divided by the average duration of the intervals. Equation (13) demonstrates the equations for rPVIs, which are calculated as the average of the duration differences between successive intervals. Further, nPVIs are the normalized values of rPVIS, as stated in Equation (14). The normalized values are expected to reduce the influence of the speech rate on raw values [37]. Rhythm features are extracted with Correlatore, a rhythmic metric analyzer [38].

$$\text{VarcoV} = \frac{\Delta\text{V} * 100}{\text{meanV}} \quad (11)$$

$$\text{VarcoC} = \frac{\Delta\text{C} * 100}{\text{meanC}} \quad (12)$$

$$\text{rPVIs} = \frac{1}{m-1}\sum_{k=1}^{m-1}|d_k - d_{k+1}| \quad (13)$$

$$\text{nPVIs} = \frac{100}{m-1}\sum_{k=1}^{m-1}\left|\frac{d_k - d_{k+1}}{\frac{d_k+d_{k+1}}{2}}\right| \quad (14)$$

## IV. Speech Datasets

Datasets from three languages, TORGO for English, QoLT for Korean, and SSNCE for Tamil, are utilized in this study. As we focus on classifying severity levels of dysarthria, we exclude healthy speakers for classification experiments. Refer to Table V for the number of speakers and utterances by languages and severity levels of dysarthria.

### *A. TORGO dataset*

Two database are commonly used for English dysarthric speech research: the UA-Speech database[39] and the TORGO database [40]. While the former database contains a larger number of speakers, it only has isolated words as materials. TORGO database, on the contrary, comprises isolated words and sentences. While isolated words may give useful information in terms of intelligibility classification, sentence stimuli have abundant acoustic information, especially related to prosody. Since this study aims to inspect acoustic measurements from different speech dimensions, we limit our analysis to the TORGO database.

TABLE V: Number of speakers and utterances

| Language | mild | | moderate | | severe | |
|---|---|---|---|---|---|---|
| | spk | utt | spk | utt | spk | utt |
| English | 2 | 219 | 2 | 87 | 4 | 107 |
| Korean | 23 | 230 | 40 | 400 | 7 | 70 |
| Tamil | 7 | 1820 | 10 | 2600 | 3 | 780 |

TORGO dataset contains 8 dysarthria speakers (5 males, 3 females). The Frenchay assessment [41], a standardized assessment for dysarthria severity classification, was performed by a speech pathologist. Consequently, the dataset contains 2 mild speakers, 1 mild-to-moderate speaker, 1 moderate-to-severe speaker, and 4 severe speakers. As mild-to-moderate and moderate-to-severe severity classes consist of a single speaker, two speakers are grouped into one moderate category. Although the TORGO dataset consists of non-words, words, restricted sentences, and spontaneous speech, we limit our experiments to sentences. A total of 413 utterances are managed in this paper.

## B. *QoLT dataset*

Quality of Life Technology (QoLT) dataset [42] is a Korean dysarthric speech corpus. The corpus collected speech from 70 dysarthric speakers (45 males, 25 females). Five speech pathologists carried out intelligibility assessments on a 5-point Likert scale. With mild-to-moderate and moderate-to-severe classes combined to moderate class for multilingual classification experiments, the dataset includes 23 mild, 40 moderate, 7 severe speakers. QoLT dataset embodies isolated words and restricted sentences, but we only use sentences. Speakers recorded five phonetically balanced sentences which are repeated twice. Consequently, a total of 700 utterances are used for the experiment.

## C. *SSNCE dataset*

SSNCE dataset [43] is a Tamil dysarthric speech corpus, publicly available by request. The dataset has 20 dysarthric speakers (13 males, 7 females). Two speech pathologists marked speech intelligibility scores for dysarthric speakers on a 7-point Likert scale. With a score of 0 considered healthy, score 1 and 2 are grouped as mild, score 3 and 4 as moderate, and score 5 and 6 as severe. Correspondingly, we used 7 mild speakers, 10 moderate speakers, and 3 severe speakers for the experiment. While the SSNCE dataset has word and sentence materials, only sentences are considered in this paper. A total of 260 unique sentences, a combination of common and uncommon Tamil phrases, were spoken by each speaker. Hence 5200 sentences are handled in this paper.

# V. Experiments

## A. *Experimental Settings*

*1) Feature selection:* In XGBoost, the feature relative importance can be calculated by several metrics, including weight, gain, and cover. Since we aim to determine the optimal feature set for each language, a gain-based importance score is utilized for feature selection. Gain measures the decrease in node impurity, meaning the improvement in performance brought by a new feature to the classifier. Therefore, the gain-based importance score is considered as the most relevant score to interpret the relative importance of features [44].

To find the best feature set for each language, classification accuracy is sequentially measured by dropping the least-important feature. Figure 1 illustrates the classification accuracy by the number of features for the Korean dataset. Starting from using all 39 acoustic features, we continue to exclude the least important feature from the feature set. As Figure 1 presents, using too large or small number of features impedes classification performances. Further information on feature importance scores by languages is summarized in Figure 2, where a vertical red line displays the criteria of the optimal feature set.

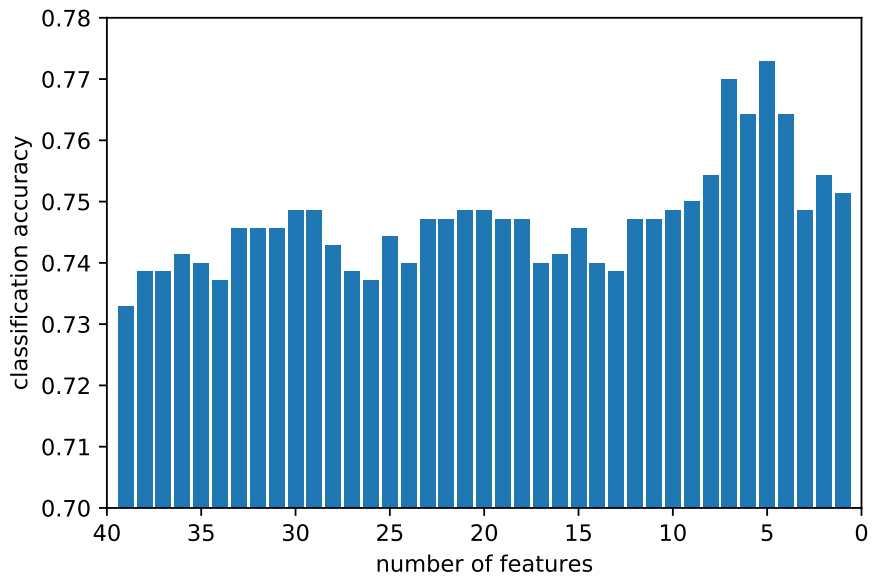

Fig. 1: Classification accuracy using different number of features

Table VI demonstrates the optimal feature sets, which are determined as the feature set with the highest classification accuracy: (**English**) HNR, PCT, min/max energy, deltaV and deltaC; (**Korean**) PCT, phone ratio, pause duration, max F0, and standard deviation energy; (**Tamil**) APQ, shimmer, degree of voice breaks, PCC, PCT, quadrilateral VSA, VAI, F2-ratio, mean/min/max/std F0, mean/min/max/std energy, VarcoV, and nPVIV. Bold indicates language-independent features across three languages, while underline refers to shared features between two languages. Accordingly, the language-independent feature set consists of PCT only. Table VII presents the part of the tabular used for our proposed method. PCT, a language-independent feature, is used for all English, Korean, and Tamil speakers. For Max F0, which is selected by Korean and Tamil, we remove max F0 values from English speakers. Lastly, as for APQ, values from English and Korean speakers are treated as missing, since it is chosen by Tamil only.

TABLE VI: Optimal feature sets

| Category | Languages | | |
|---|---|---|---|
| | English | Korean | Tamil |
| voice quality | HNR | - | APQ, shimmer, degree of voice breaks |
| phoneme correctness | **PCT** | **PCT** | PCC, **PCT** |
| vowel distortion | - | - | VSA□, VAI, F2-ratio |
| speech rate | - | pause duration, phone ratio | - |
| pitch | - | max F0 | mean/min/max/std F0 |
| loudness | min/max energy | std energy | mean/min/max/std energy |
| rhythm | deltaV, deltaC | - | VarcoV, nPVIV |

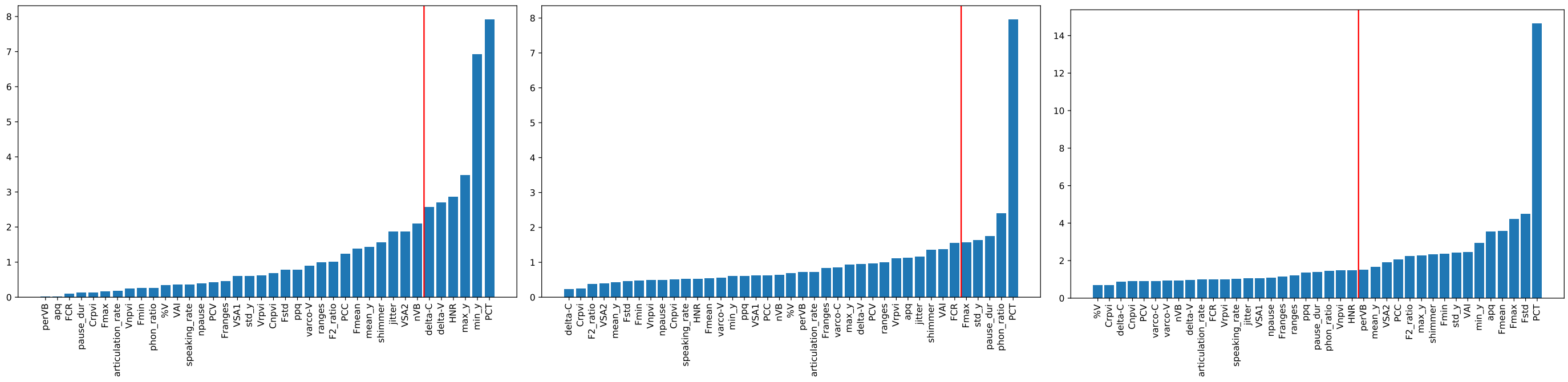

(a) English (b) Korean (c) Tamil

Fig. 2: Gain-based importance scores

TABLE VII: Tabular for the proposed method

| Language | PCT | max F0 | APQ |
|---|---|---|---|
| English | O | N/A | N/A |
| Korean | O | O | N/A |
| Tamil | O | O | O |

Monolingual classifications using the optimal feature set are further conducted, to verify the efficacy of the selected feature set. Table VIII presents the F1 scores of monolingual classifications by languages. Experimental results indicate that 39 acoustic features are beneficial for dysarthria severity classifications. While English, Korean, and Tamil each achieved 36.96%, 62.71%, and 40.36% F1-score with MFCCs, the performance enhanced to 48.07%, 72.64%, and 46.52%, respectively, when all 39 acoustic features are included. Additionally, further performance enhancements are attained with feature selection, to the F1-score of 59.03%, 76.40%, and 46.91% for English, Korean, and Tamil, respectively. The results prove the effectiveness of the optimal feature set for each language.

TABLE VIII: Monolingual classification (F1-score)

| Features | English | Korean | Tamil | Average |
|---|---|---|---|---|
| MFCCs | 36.96 | 62.71 | 40.36 | 46.68 |
| MFCCs + All | 48.07 | 72.64 | 46.52 | 55.74 |
| **Selected (gain)** | **59.03** | **76.40** | **46.91** | **60.78** |

*2) Classification:* Classification is conducted by using XGBoost algorithm. Each classifier is optimized via grid search. Because we use large enough number of boost rounds/estimators (100 to 1000), max depth is fixed to 3, while default settings are used for other hyper-parameters. Lastly, Leave-One-Subject-Out Cross Validation (LOSOCV) approach is utilized, for better generalizability of the results with imbalanced dataset.

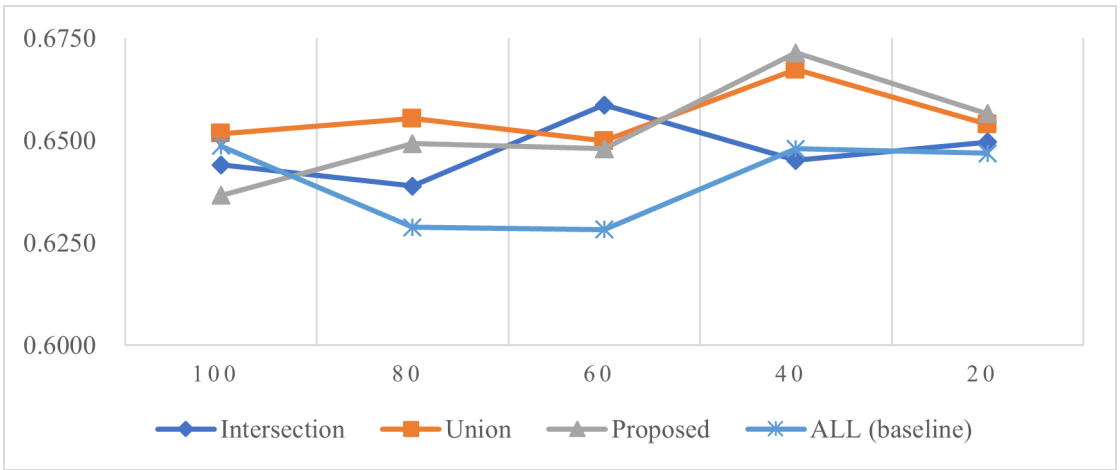

Fig. 3: Average F1-score with different sizes of Tamil dataset

Overfitting to the largest dataset, the Tamil dataset, can worsen the classification performances of other smaller English and Korean datasets. Experiments using different sizes of Tamil dataset are implemented to alleviate this problem. Utterances are randomly selected among 260 unique sentences - 208 sentences for 80%, 156 sentences for 60%, 104 sentences for 40%, and 52 sentences for 20%. Figure 3 exhibits the average F1-score of the three languages by using different sizes of Tamil dataset. Compared to cross-lingual classification using MFCCs and all 39 acoustic features, experiments using the selected feature set generally show better performances. Furthermore, when the performances of the intersection, the union, and the proposed method are compared, overall performance of experiments using the intersection feature set is the lowest. Moreover, while the union experiment achieves higher F1 scores with large datasets, the proposed experiment exceeds union experiment as the dataset gets smaller. This implies that although XGBoost classifiers can make proper decisions on feature selection with large enough dataset, explicit selection for language-sensitive features is required with smaller dataset. The best average F1 score is achieved when using 40% of the Tamil dataset. Accordingly, further experiments are conducted using this setting.

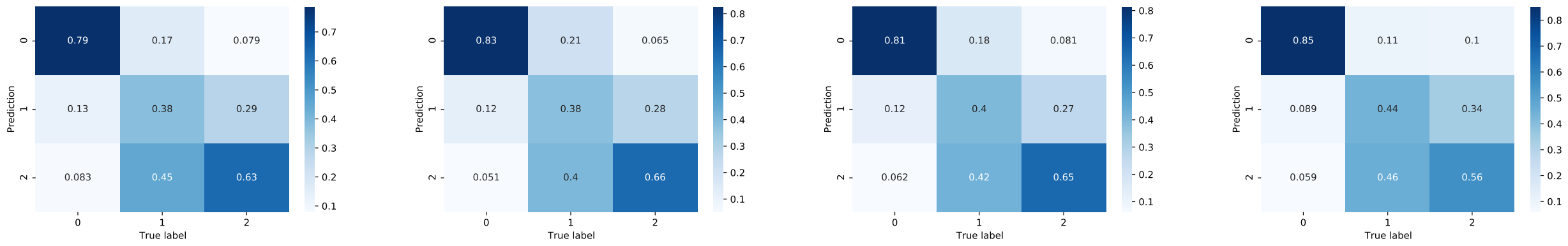

(a) English˙All (b) English˙Intersection (c) English˙Union (d) English˙Proposed

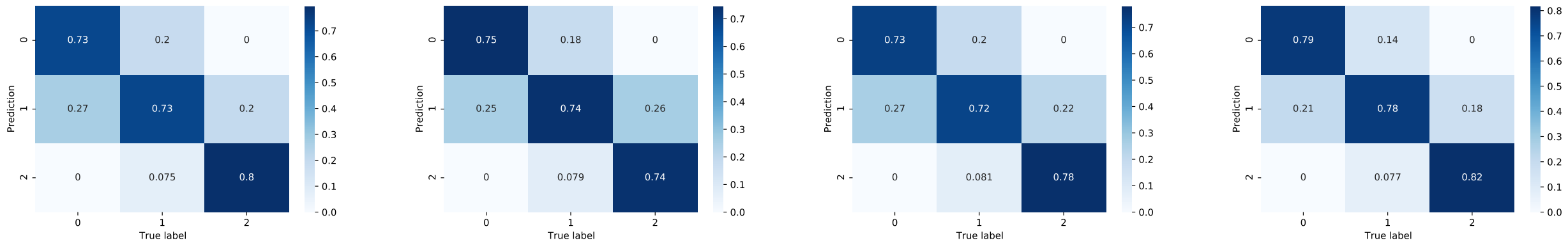

(e) Korean˙All (f) Korean˙Intersection (g) Korean˙Union (h) Korean˙Proposed

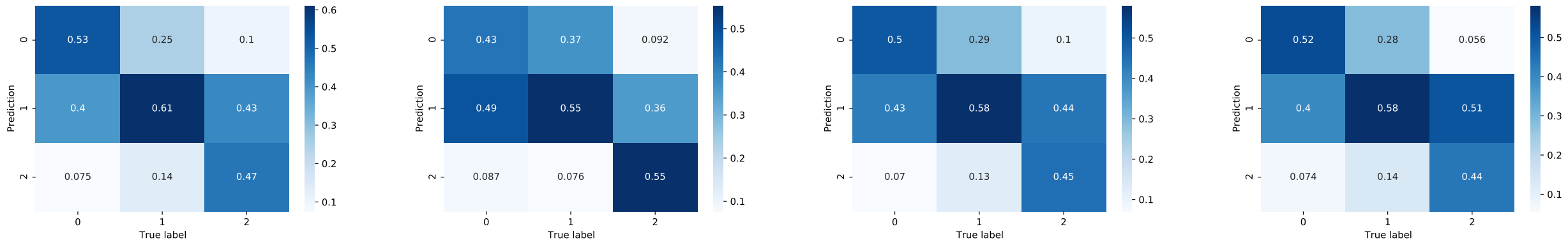

(i) Tamil˙All (j) Tamil˙Intersection (k) Tamil˙Union (l) Tamil˙Proposed

Fig. 4: Confusion Matrix for MFCCs+ALL, Intersection, Union, and Proposed experiments

### *B. Experimental results*

*1) Cross-lingual classification:* Table IX demonstrates the F1-score of multilingual classification experiments, with the best performances indicated in bold. The proposed method performs best in terms of the average value of the three languages, attaining 67.14% F1-score. Union experiment comes after the proposed method with 66.74%, followed by MFCCs+ALL experiment of 65.07%. The intersection experiment shows worse performances than using all features, with 64.52% F1-score. Experimental results imply that our proposed method shows robust performances across languages with different prosodic systems. The result also indicates that different types of features should be considered for cross-lingual classifications.

TABLE IX: Cross-lingual classification (F1-score)

| Experiment | English | Korean | Tamil | Average |
|---|---|---|---|---|
| MFCCs | 60.66 | 59.64 | 38.78 | 53.03 |
| MFCCs + All | 66.99 | 72.47 | **55.75** | 65.07 |
| Intersection | 69.14 | 73.56 | 50.85 | 64.52 |
| Union | **71.96** | 72.76 | 55.48 | 66.74 |
| **Proposed** | 69.46 | **78.13** | 53.84 | **67.14** |

For English datasets, the union experiment shows the highest F1-score for severity classification with 71.96%, with the intersection of 69.14% and the proposed method of 69.46%. As for Korean, the proposed method attains the highest F1-score of 78.13%, contrasted to 73.56% for intersection and 72.76% for the union experiment. On the other hand, Tamil shows the best performance when using all features from all three languages achieving 55.75% F1-score, compared to 50.85% for intersection, 55.48% for union, and 53.84% for the proposed method. This result aligns with the results from Figure 3, where the classifier can make proper decisions without explicit feature selection, with large enough datasets. Figure 4 presents the confusion matrix for MFCC+ALL, intersection, union, and proposed experiments. Score 0, 1, 2 each refers to mild, moderate, and severe class. With analysis of classification results by severity levels, the proposed model achieves better performances especially in discriminating mild and moderate speakers. However, the capability of distinguishing moderate and severe speakers is lower than other approaches, for English and Tamil. We plan to mitigate these weaknesses in future studies.

### *C. Comparisons with Mono-lingual classifier*

Table X presents the relative percentage increase from monolingual classifiers to the proposed method. The proposed method shows a relative percentage increase of 17.67%, 2.28%, and 7.79% for English, Korean, and Tamil, respectively, in contrast to monolingual classifiers. This validates the efficacy of our proposed method, which aims to benefit from augmented data by combining the three datasets from different languages.

TABLE X: Comparison with mono-lingual classifiers

| Language | Mono-lingual | Proposed | Relative Increase |
|---|---|---|---|
| English | 59.03 | **69.46** | 17.67 |
| Korean | 76.39 | **78.13** | 2.28 |
| Tamil | 49.95 | **53.84** | 7.79 |
| Average | 61.79 | **67.14** | 8.66 |

## VI. Conclusion

This paper specifies that commonly shared features and language-specific features must be considered for cross-language dysarthria severity classification. We propose a method of cross-lingual dysarthria severity classification utilizing both shared and unique feature sets. Especially, different language-sensitive features are removed and treated as missing values depending on the target language. According to the experimental results, the proposed method achieves the best average performance of the three languages- English, Korean, and Tamil, compared to baseline experiments using the intersection set or the union set of the three optimal feature sets. The result indicates that the proposed method shows robust performances even with languages with different prosodic systems. Further, our proposed method exceeded mono-lingual classifiers for all three languages, benefiting from larger data for training. To conclude, the proposed method benefits from not only the augmented data by shared features across languages but additional language-unique features, designed to prevent interference of unhelpful features from other languages.

A limitation of this study lies in the potential problem that XGBoost takes the default direction for the missing values. While our proposed method assures the best possible result with former choices, it does not guarantee that those previous choices, which may contain a default direction, are the best ones on the whole [23]. This implies that the more missing values, the higher the probability that the model may make wrong choices. Hence, with our proposed model using different numbers of features, English, which drops the most number of features among the three languages, may have suffered from such problems. Our future research aims to mitigate such problems. The future study also includes using different datasets and languages to validate the generalizability of our proposed model. Employing deep neural networks which are designed to exploit both language-independent and language-sensitive features can be another solution for cross-lingual classification to be discovered. Lastly, while this study could not consider dysarthric types due to limited access to data, considerations of dysarthric types are also necessary.

## Acknowledgment

This work was supported by Institute of Information & communications Technology Planning & Evaluation (IITP) grant funded by the Korea government(MSIT) (No.2022-0-00223, Development of digital therapeutics to improve communication ability of autism spectrum disorder patients)

## References

[1] P Enderby, "Disorders of communication: dysarthria," *Handbook of clinical neurology*, vol. 110, pp. 273–281, 2013.

[2] Myung Jong Kim and Hoirin Kim, "Combination of multiple speech dimensions for automatic assessment of dysarthric speech intelligibility," in *Thirteenth Annual Conference of the International Speech Communication Association*, 2012.

[3] EJ Yeo, S Kim, and M Chung, "Automatic severity classification of korean dysarthric speech using phoneme-level pronunciation features.," in *Interspeech*, 2021, pp. 4838–4842.

[4] Feifei Xiong, Jon Barker, Zhengjun Yue, and Heidi Christensen, "Source domain data selection for improved transfer learning targeting dysarthric speech recognition," in *ICASSP 2020-2020 IEEE International Conference on Acoustics, Speech and Signal Processing (ICASSP)*. IEEE, 2020, pp. 7424–7428.

[5] SR Mani Sekhar, Gaurav Kashyap, Akshay Bhansali, Kushan Singh, et al., "Dysarthric-speech detection using transfer learning with convolutional neural networks," *ICT Express*, 2021.

[6] Pratibha Dumane, Bilal Hungund, and Satishkumar Chavan, "Dysarthria detection using convolutional neural network," in *Techno-Societal 2020*, pp. 449–457. Springer, 2021.

[7] Amlu Anna Joshy and Rajeev Rajan, "Automated dysarthria severity classification using deep learning frameworks," in *2020 28th European Signal Processing Conference (EUSIPCO)*. IEEE, 2021, pp. 116–120.

[8] John Harvill, Dias Issa, Mark Hasegawa-Johnson, and Changdong Yoo, "Synthesis of new words for improved dysarthric speech recognition on an expanded vocabulary," in *ICASSP 2021-2021 IEEE International Conference on Acoustics, Speech and Signal Processing (ICASSP)*. IEEE, 2021, pp. 6428–6432.

[9] Yishan Jiao, Ming Tu, Visar Berisha, and Julie Liss, "Simulating dysarthric speech for training data augmentation in clinical speech applications," in *2018 IEEE international conference on acoustics, speech and signal processing (ICASSP)*. IEEE, 2018, pp. 6009–6013.

[10] Zengrui Jin, Mengzhe Geng, Xurong Xie, Jianwei Yu, Shansong Liu, Xunying Liu, and Helen Meng, "Adversarial data augmentation for disordered speech recognition," *arXiv preprint arXiv:2108.00899*, 2021.

[11] Mengzhe Geng, Xurong Xie, Shansong Liu, Jianwei Yu, Shoukang Hu, Xunying Liu, and Helen Meng, "Investigation of data augmentation techniques for disordered speech recognition," *arXiv preprint arXiv:2201.05562*, 2022.

[12] TA Mariya Celin, T Nagarajan, and P Vijayalakshmi, "Data augmentation using virtual microphone array synthesis and multi-resolution feature extraction for isolated word dysarthric speech recognition," *IEEE Journal of Selected Topics in Signal Processing*, vol. 14, no. 2, pp. 346–354, 2020.

[13] Y Kim and Y Choi, "A cross-language study of acoustic predictors of speech intelligibility in individuals with parkinson's disease," *Journal of Speech, Language, and Hearing Research*, vol. 60, no. 9, pp. 2506–2518, 2017.

[14] L Moro-Velazquez, JA Gomez-Garcia, J Godino-Llorente, F Grandas-Perez, S Shattuck-Hufnagel, V Yagüe-Jimenez, and N9 Dehak, "Phonetic relevance and phonemic grouping of speech in the automatic detection of parkinson's disease," *Scientific reports*, vol. 9, no. 1, pp. 1–16, 2019.

[15] D Kovác, "Multilingual analysis of hypokinetic dysarthria in patients with parkinson's disease," .

[16] D Kovac, J Mekyska, Z Galaz, L Brabenec, M Kostalova, SZ Rapcsak, and I Rektorova, "Multilingual analysis of speech and voice disorders in patients with parkinson's disease," in *44th International Conference on TSP*. IEEE, 2021, pp. 273–277.

[17] Tianqi Chen and Carlos Guestrin, "XGBoost: A scalable tree boosting system," in *Proceedings of the 22nd ACM SIGKDD International Conference on Knowledge Discovery and Data Mining*, New York, NY, USA, 2016, KDD '16, pp. 785–794, ACM.

[18] Adeola Ogunleye and Qing-Guo Wang, "Xgboost model for chronic kidney disease diagnosis," *IEEE/ACM transactions on computational biology and bioinformatics*, vol. 17, no. 6, pp. 2131–2140, 2019.

[19] Junchen Bao, "Multi-features based arrhythmia diagnosis algorithm using xgboost," in *2020 International Conference on Computing and Data Science (CDS)*. IEEE, 2020, pp. 454–457.

[20] Meihong Ma, Gang Zhao, Bingshun He, Qing Li, Haoyue Dong, Shenggang Wang, and Zhongliang Wang, "Xgboost-based method for flash flood risk assessment," *Journal of Hydrology*, vol. 598, pp. 126382, 2021.

[21] Maria Athanasiou, Konstantina Sfrintzeri, Konstantia Zarkogianni, Anastasia C Thanopoulou, and Konstantina S Nikita, "An explainable xgboost–based approach towards assessing the risk of cardiovascular disease in patients with type 2 diabetes mellitus," in *2020 IEEE 20th International Conference on Bioinformatics and Bioengineering (BIBE)*. IEEE, 2020, pp. 859–864.

[22] NVIDIA, "Xgboost," https://www.nvidia.com/en-us/glossary/data-science/xgboost.

[23] Massimo Belloni, "Xgboost is not black magic," https://towardsdatascience.com/xgboost-is-not-black-magic-56ca013144b4.

[24] Brian McFee, Colin Raffel, Dawen Liang, Daniel P Ellis, Matt McVicar, Eric Battenberg, and Oriol Nieto, "librosa: Audio and music signal analysis in python," in *Proceedings of the 14th python in science conference*. Citeseer, 2015, vol. 8, pp. 18–25.

[25] T Schölderle, A Staiger, R Lampe, K Strecker, and W Ziegler, "Dysarthria in adults with cerebral palsy: Clinical presentation and impacts on communication," *Journal of Speech, Language, and Hearing Research*, vol. 59, no. 2, pp. 216–229, 2016.

[26] A Hernandez, S Kim, and M Chung, "Prosody-based measures for automatic severity assessment of dysarthric speech," *Applied Sciences*, vol. 10, no. 19, pp. 6999, 2020.

[27] Y Koike, "Application of some acoustic measures for the evaluation of laryngeal dysfunction.," *Studia Phonetica*, vol. 7, pp. 17–23, 1973.

[28] P Boersma, "Praat: doing phonetics by computer [computer program]," *http://www. praat. org/*, 2011.

[29] LJ Platt, G Andrews, and PM Howie, "Dysarthria of adult cerebral palsy: Ii. phonemic analysis of articulation errors," *Journal of Speech, Language, and Hearing Research*, vol. 23, no. 1, pp. 41–55, 1980.

[30] J Grosman, "Fine-tuned XLSR-53 large model for speech recognition in English," https://huggingface.co/jonatasgrosman/wav2vec2-large-xlsr-53-english, 2021.

[31] EJ Yeo, "wav2vec2-large-xlsr-korean," https://huggingface.co/speech31/xlsr-large-korean-ksponspeech, 2022.

[32] P von Patrick, "wav2vec2-large-xlsr-53-tamil," https://huggingface.co/Amrrs/wav2vec2-large-xlsr-53-tamil, 2020.

[33] KL Lansford and JM Liss, "Vowel acoustics in dysarthria: speech disorder diagnosis and classification," *Journal of Speech, Language, and Hearing Research*, vol. 57, no. 1, pp. 57–68, 2014.

[34] M McAuliffe, M Socolof, S Mihuc, M Wagner, and M Sonderegger, "Montreal forced aligner: Trainable text-speech alignment using kaldi.," in *Interspeech*, 2017, vol. 2017, pp. 498–502.

[35] Y Jadoul, B Thompson, and BD Boer, "Introducing parselmouth: A python interface to praat," *Journal of Phonetics*, vol. 71, pp. 1–15, 2018.

[36] Abner Hernandez, Eun Jung Yeo, Sunhee Kim, and Minhwa Chung, "Dysarthria detection and severity assessment using rhythm-based metrics.," in *INTERSPEECH*, 2020, pp. 2897–2901.

[37] V Dellwo and P Wagner, "Relationships between speech rate and rhythm," in *Proceedings of the ICPhS*, 2003.

[38] P Mairano and A Romano, "Un confronto tra diverse metriche ritmiche usando correlatore," *La dimensione temporale del parlato*, vol. 5, 2010.

[39] Heejin Kim, Mark Hasegawa-Johnson, Adrienne Perlman, Jon Gunderson, Thomas S Huang, Kenneth Watkin, and Simone Frame, "Dysarthric speech database for universal access research," in *Ninth Annual Conference of the International Speech Communication Association*, 2008.

[40] F Rudzicz, AK Namasivayam, and T Wolff, "The torgo database of acoustic and articulatory speech from speakers with dysarthria," *Language Resources and Evaluation*, vol. 46, no. 4, pp. 523–541, 2012.

[41] P Enderby, "Frenchay dysarthria assessment," *British Journal of Disorders of Communication*, vol. 15, no. 3, pp. 165–173, 1980.

[42] D Choi, B Kim, Y Kim, Y Lee, Y Um, and M Chung, "Dysarthric speech database for development of qolt software technology," in *LREC*, 2012, pp. 3378–3381.

[43] MC TA, T Nagarajan, and P Vijayalakshmi, "Dysarthric speech corpus in tamil for rehabilitation research," in *Region TENCON*. IEEE, 2016, pp. 2610–2613.

[44] Chunyu Zhang, Danshi Wang, Chuang Song, Lingling Wang, Jianan Song, Luyao Guan, and Min Zhang, "Interpretable learning algorithm based on xgboost for fault prediction in optical network," in *2020 Optical Fiber Communications Conference and Exhibition (OFC)*. IEEE, 2020, pp. 1–3.